\documentclass{article}

\usepackage{microtype}
\usepackage{graphicx}
\usepackage{booktabs}
\usepackage{amsmath,amssymb,mathtools,amsthm}
\usepackage{multirow}
\usepackage{tabularx}
\usepackage{colortbl}
\usepackage{xcolor}
\usepackage{algorithm}
\usepackage{algorithmic}

\definecolor{claudec}{HTML}{1F77B4}
\definecolor{geminic}{HTML}{2CA02C}
\definecolor{deepseekc}{HTML}{D62728}
\definecolor{gptc}{HTML}{9467BD}
\definecolor{llamac}{HTML}{FF7F0E}

\definecolor{accentnavy}{HTML}{0B3D91}
\definecolor{accentteal}{HTML}{17A2A2}
\definecolor{accentamber}{HTML}{E08E0B}
\definecolor{softbg}{HTML}{F5F7FA}
\definecolor{softborder}{HTML}{C9D2DC}

\usepackage{tikz}
\usepackage{pgfplots}
\pgfplotsset{compat=1.17}
\usetikzlibrary{shapes.geometric,arrows.meta,positioning,fit,calc,
  backgrounds,decorations.pathreplacing,shapes,matrix,patterns,shadows}

\usepackage[round,numbers]{natbib}
\usepackage{hyperref}
\hypersetup{colorlinks=true,citecolor=accentnavy,linkcolor=accentnavy,urlcolor=accentnavy}

\usepackage[accepted]{icml2026}

\newtheorem{assumption}{Assumption}
\newtheorem{remark}{Remark}

\newcommand{\CCB}{\textsc{CCB}}
\newcommand{\pd}{p_d}
\newcommand{\kstar}{k^*}

\icmltitlerunning{The Complexity Ceiling Benchmark}

\begin{document}

\twocolumn[
\icmltitle{The Complexity Ceiling Benchmark: \\A Multi-Domain Evaluation of Sequential Reasoning Under Depth Scaling}

\begin{icmlauthorlist}
\icmlauthor{Shubh Chapra}{1}
\icmlauthor{Dhruv Kumar}{1}
\icmlauthor{Murari Mandal}{2}
\icmlauthor{Yash Sinha}{1}
\end{icmlauthorlist}

\icmlaffiliation{1}{BITS Pilani, Pilani Campus}
\icmlaffiliation{2}{KIIT Bhubaneswar}
\icmlcorrespondingauthor{}{f20230005@pilani.bits-pilani.ac.in}
\icmlkeywords{Reasoning, Benchmarks, LLMs, Depth Scaling, Geometric Decay}

\vskip 0.3in
]

\printAffiliationsAndNotice{}

\begin{abstract}
We introduce the Complexity Ceiling Benchmark (\CCB), a controlled
evaluation of how language-model reasoning decays as the number of
required sequential steps grows. \CCB{} fixes the semantic content of a
task and varies only its depth $N{\in}\{5,\dots,50\}$ across three
structurally distinct regimes: grounded spatial state-tracking,
abstract symbolic pointer manipulation, and transitive relational
inference. Across 6{,}000 trials over five frontier and open-weight
LLMs we find a consistent pattern of geometric per-step decay with
widely separated domain ceilings: on the first two regimes the strongest
models retain $\pd>0.92$ across $N{=}50$; on the third every model
collapses by $N{=}5$, with the best model's 50\%-success horizon at
$H_{0.5}{\approx}4.7$ steps despite $\pd{=}0.863$. A trace-level metric
(\textsc{TFBC}) shows that $14.5\%$ of correct answers across the
benchmark are reached via incorrect intermediate reasoning. Forced verbose state-tracking does not
move the ceiling (McNemar $p{=}1.000$), and the mean step at which
reasoning first diverges, $\kstar$, predicts within-domain accuracy
better than parameter count. \CCB{} and the geometric decay model
together reduce a model's long-horizon reasoning profile to one
interpretable number per task family.
\end{abstract}

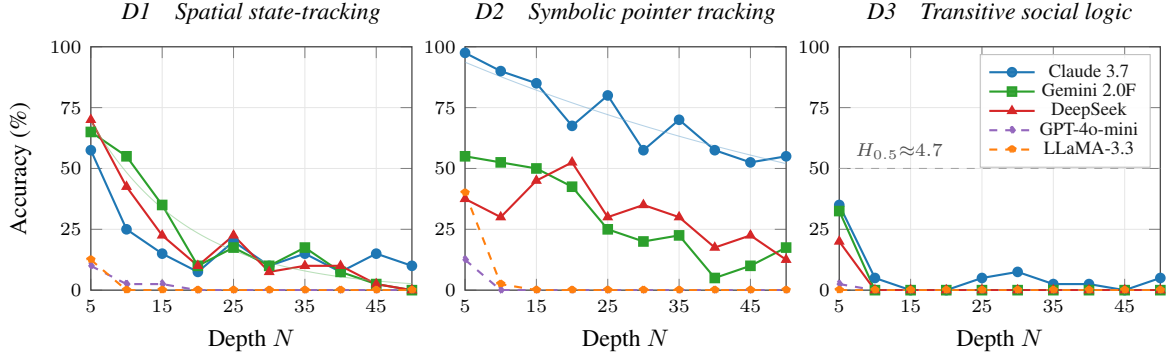
\begin{figure*}[!tb]
\centering
\begin{tikzpicture}
\pgfplotsset{
  herodecay/.style={
    width=0.34\textwidth, height=4.8cm,
    xlabel={Depth $N$}, ymin=0, ymax=100, xmin=5, xmax=50,
    ytick={0,25,50,75,100}, xtick={5,15,25,35,45},
    tick label style={font=\scriptsize},
    label style={font=\small},
    title style={font=\small\itshape, yshift=-1pt},
    grid=major, grid style={line width=0.1pt,draw=gray!20},
    axis line style={draw=black!70, line width=0.5pt},
    every axis plot/.append style={thick}
  }
}

\begin{axis}[herodecay, title={D1\quad Spatial state-tracking},
  ylabel={Accuracy (\%)}, name=axA]
\addplot[claudec,mark=*,mark size=1.6pt] coordinates {(5,57.5)(10,25)(15,15)(20,7.5)(25,20)(30,10)(35,15)(40,7.5)(45,15)(50,10)};
\addplot[geminic,mark=square*,mark size=1.6pt] coordinates {(5,65)(10,55)(15,35)(20,10)(25,17.5)(30,10)(35,17.5)(40,7.5)(45,2.5)(50,0)};
\addplot[deepseekc,mark=triangle*,mark size=1.6pt] coordinates {(5,70)(10,42.5)(15,22.5)(20,10)(25,22.5)(30,7.5)(35,10)(40,10)(45,2.5)(50,0)};
\addplot[gptc,mark=diamond*,mark size=1.4pt,dashed] coordinates {(5,10)(10,2.5)(15,2.5)(20,0)(25,0)(30,0)(35,0)(40,0)(45,0)(50,0)};
\addplot[llamac,mark=pentagon*,mark size=1.4pt,dashed] coordinates {(5,12.5)(10,0)(15,0)(20,0)(25,0)(30,0)(35,0)(40,0)(45,0)(50,0)};
\addplot[geminic!60,thin,domain=5:50,samples=46,smooth,opacity=0.55]
  {100 * 0.930^x};
\end{axis}

\begin{axis}[herodecay, title={D2\quad Symbolic pointer tracking},
  at={(axA.south east)}, xshift=2.0em, anchor=south west, name=axB]
\addplot[claudec,mark=*,mark size=1.6pt] coordinates {(5,97.5)(10,90)(15,85)(20,67.5)(25,80)(30,57.5)(35,70)(40,57.5)(45,52.5)(50,55)};
\addplot[geminic,mark=square*,mark size=1.6pt] coordinates {(5,55)(10,52.5)(15,50)(20,42.5)(25,25)(30,20)(35,22.5)(40,5)(45,10)(50,17.5)};
\addplot[deepseekc,mark=triangle*,mark size=1.6pt] coordinates {(5,37.5)(10,30)(15,45)(20,52.5)(25,30)(30,35)(35,30)(40,17.5)(45,22.5)(50,12.5)};
\addplot[gptc,mark=diamond*,mark size=1.4pt,dashed] coordinates {(5,12.5)(10,0)(15,0)(20,0)(25,0)(30,0)(35,0)(40,0)(45,0)(50,0)};
\addplot[llamac,mark=pentagon*,mark size=1.4pt,dashed] coordinates {(5,40)(10,2.5)(15,0)(20,0)(25,0)(30,0)(35,0)(40,0)(45,0)(50,0)};
\addplot[claudec!60,thin,domain=5:50,samples=46,smooth,opacity=0.55]
  {100 * 0.987^x};
\end{axis}

\begin{axis}[herodecay, title={D3\quad Transitive social logic},
  at={(axB.south east)}, xshift=2.0em, anchor=south west, name=axC,
  legend style={font=\scriptsize, at={(0.97,0.97)}, anchor=north east,
    draw=black!25, fill=white, fill opacity=0.92,
    row sep=-2pt, nodes={inner sep=1.5pt}}]
\addplot[claudec,mark=*,mark size=1.6pt] coordinates {(5,35)(10,5)(15,0)(20,0)(25,5)(30,7.5)(35,2.5)(40,2.5)(45,0)(50,5)};
\addlegendentry{Claude 3.7}
\addplot[geminic,mark=square*,mark size=1.6pt] coordinates {(5,32.5)(10,0)(15,0)(20,0)(25,0)(30,0)(35,0)(40,0)(45,0)(50,0)};
\addlegendentry{Gemini 2.0F}
\addplot[deepseekc,mark=triangle*,mark size=1.6pt] coordinates {(5,20)(10,0)(15,0)(20,0)(25,0)(30,0)(35,0)(40,0)(45,0)(50,0)};
\addlegendentry{DeepSeek}
\addplot[gptc,mark=diamond*,mark size=1.4pt,dashed] coordinates {(5,2.5)(10,0)(15,0)(20,0)(25,0)(30,0)(35,0)(40,0)(45,0)(50,0)};
\addlegendentry{GPT-4o-mini}
\addplot[llamac,mark=pentagon*,mark size=1.4pt,dashed] coordinates {(5,0)(10,0)(15,0)(20,0)(25,0)(30,0)(35,0)(40,0)(45,0)(50,0)};
\addlegendentry{LLaMA-3.3}
\draw[gray!70,dashed,thin] (axis cs:5,50) -- (axis cs:50,50);
\node[font=\scriptsize\itshape,text=gray!60!black,anchor=west]
  at (axis cs:6,57) {$H_{0.5}{\approx}4.7$};
\end{axis}

\end{tikzpicture}
\caption{\textbf{The Complexity Ceiling.}
  Accuracy as a function of depth $N$ across three structurally
  distinct reasoning regimes, with semantic content held fixed.
  Markers: empirical accuracy at $n{=}40$ trials per cell. Thin solid
  curves in D1 and D2: fitted geometric model $100{\cdot}\pd^N$ for
  the top frontier model (Gemini in D1, Claude in D2). On
  D1 and D2 frontier models track the geometric decay with
  $\pd{\in}[0.92,0.99]$, leaving meaningful accuracy at $N{=}50$. On
  transitive social logic (D3) every model collapses past $N{=}5$
  regardless of capability tier; even the best model's $50\%$-success
  horizon is $H_{0.5}{\approx}4.7$ steps. Solid lines: frontier or
  competitive open-weight; dashed: smaller models. $6{,}000$ trials
  total.}
\label{fig:hero}
\end{figure*}

\section{Introduction}

When a language model fails on a long reasoning task, current benchmarks
cannot tell us why. The model may lack knowledge, misread the prompt,
exhaust its context window, or simply lose its place across the
intermediate steps. These failure modes have different remedies and
different implications for deploying LLMs as agents, but aggregate
accuracy on a fixed-difficulty benchmark sees only the union: a
single number that says the model got it wrong. The most consequential
of these modes for agentic use;  the cumulative loss of coherence as
the number of sequential steps grows;  is also the one least visible
to such benchmarks \citep{dziri2023faith,sinha2025illusion}.

This paper treats depth as the controlled experimental variable.
Figure~\ref{fig:hero} shows the central result of doing so. The
Complexity Ceiling Benchmark (\CCB) holds task semantics fixed and
sweeps the number of required reasoning steps $N$ from 5 to 50 across
three structurally distinct regimes: grounded spatial state-tracking
(D1), abstract symbolic pointer manipulation (D2), and transitive
relational inference (D3). Six thousand trials over five frontier and
open-weight LLMs reveal a consistent pattern;  per-step retention
decays geometrically in $N$;  and a striking dissociation between
domains. On D1 and D2 the strongest models hold retention probability
$\pd>0.92$ across the full depth range, leaving meaningful accuracy at
$N{=}50$. On D3, every model collapses past $N{=}5$, regardless of
capability tier; the best evaluated model's 50\%-success horizon is
only $H_{0.5}{\approx}4.7$ steps. Tracing the intermediate reasoning
adds a second result that aggregate accuracy hides: across the
benchmark, $14.5\%$ of correct answers come from traces that
diverged from the canonical reasoning, and the share is highest on the
hardest domain.

\paragraph{Contributions.}
The geometric decay model and the trace-level metric we introduce
together let us summarise a (model, task-family) pair with two
numbers: a per-step retention probability $\pd$ and a mean trace-divergence
step $\kstar$. Both connect to deployment: $\pd$ feeds the horizon-length
framework of \citet{sinha2025illusion} and determines the depth at which
sustained accuracy crosses any chosen threshold, and $\kstar$ predicts
within-domain accuracy more faithfully than parameter count. The
findings in this paper are scoped to vanilla autoregressive inference
on the five evaluated models; evaluating process-supervised and
tool-augmented systems is the obvious next step and is one of the uses
\CCB{} is built to serve.

\section{Related Work}
\label{sec:related}

\paragraph{Depth-scaling and compositional generalisation.}
Most prior work treats reasoning failure as an aggregate property of a
task. SCAN \citep{lake2018scan} and BIG-Bench
\citep{srivastava2022beyond} probe compositional generalisation at
roughly constant difficulty, providing a strong test of out-of-distribution
generalisation but limited leverage for a scaling analysis. CLUTRR
\citep{sinha2019clutrr} comes closest to our setting;  multi-hop
relational reasoning over kinship graphs;  and motivates D3, but
gives only final-answer judgements and a small fixed range of hops.
\citet{dziri2023faith} showed transformers unroll memorised subgraphs
with catastrophic failure at compositional out-of-distribution depths;
\citet{hou2026wmfam} tied this to cumulative state-tracking load;
SokoBench \citep{sokobench2026} isolates planning depth in Sokoban,
TopoBench \citep{topobench2026} shows structured state aids reasoning.
\CCB{} differs from all of these by making depth a continuous
parametric axis across three heterogeneous domains and by recovering
a single one-parameter summary statistic that lets us compare
\emph{how} a model fails, not just whether it does.

\paragraph{Trace-level evaluation and structural uncertainty.}
A parallel line of work evaluates the reasoning chain itself rather
than the answer. ROSCOE \citep{golovneva2022roscoe}, ReCEval
\citep{prasad2023receval}, and MME-CoT \citep{jiang2025mmecot}
introduce trace-quality metrics, but mostly rely on LLM-as-judge or
learned rubrics whose own reliability is contested
\citep{chaudhury2026structural}. \CCB{} sidesteps this by comparing
against a deterministically generated canonical trace; the
\textsc{TFBC} metric below requires no external scorer.

\paragraph{Process supervision and horizon connection.}
Finally, our $\pd^N$ model is the empirical counterpart of the
horizon-length framework of \citet{sinha2025illusion}, which derives an
effective task horizon $H_s{\approx}\ln(s)/\ln(\pd)$ from per-step
accuracy. \citet{cobbe2021training} showed that process-level
supervision shifts the relevant quantity; recursive scaffolds
\citep{yang2026recursive} and fast-slow recurrence
\citep{takashiro2026thinking} target the same bottleneck through
architecture. \citet{kim2024token} argue that autoregressive token
ordering is itself an inductive bias on what reasoning patterns are
accessible. Extended discussion appears in
Appendix~\ref{app:related_extended}.

\section{Benchmark Design}
\label{sec:framework}

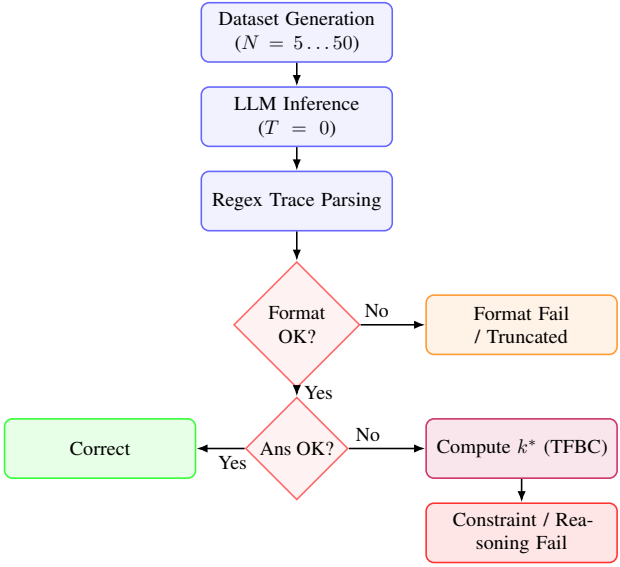
\begin{figure}[t]
\centering
\resizebox{\linewidth}{!}{
\begin{tikzpicture}[node distance=1.5cm, auto, thick,
  every node/.style={transform shape}]
\tikzstyle{block} = [rectangle, draw=blue!60, fill=blue!5, text width=9em,
  text centered, rounded corners, minimum height=3em, thick]
\tikzstyle{line} = [draw, -{Latex[length=2mm]}, thick]
\tikzstyle{decision} = [diamond, draw=red!60, fill=red!5, text width=4.5em,
  text badly centered, inner sep=0pt, thick]
\node [block] (gen) {Dataset Generation\\($N=5 \dots 50$)};
\node [block, below of=gen] (model) {LLM Inference\\($T=0$)};
\node [block, below of=model] (parse) {Regex Trace Parsing};
\node [decision, below of=parse, node distance=2.2cm] (fmt) {Format OK?};
\node [block, right of=fmt, node distance=4cm, draw=orange!80,
  fill=orange!10] (fail) {Format Fail / Truncated};
\node [decision, below of=fmt, node distance=2.2cm] (ans) {Ans OK?};
\node [block, left of=ans, node distance=3.5cm, draw=green!80,
  fill=green!10] (corr) {Correct};
\node [block, right of=ans, node distance=4cm, draw=purple!80,
  fill=purple!10] (div) {Compute $k^*$ (TFBC)};
\node [block, below of=div, draw=red!80, fill=red!10] (reason)
  {Constraint / Reasoning Fail};
\path [line] (gen) -- (model);
\path [line] (model) -- (parse);
\path [line] (parse) -- (fmt);
\path [line] (fmt) -- node [near start] {No} (fail);
\path [line] (fmt) -- node {Yes} (ans);
\path [line] (ans) -- node [near start] {Yes} (corr);
\path [line] (ans) -- node [near start] {No} (div);
\path [line] (div) -- (reason);
\end{tikzpicture}
}
\caption{\textbf{The CCB evaluation pipeline.} LLM outputs are routed through a strict
  parsing hierarchy to prevent confounding reasoning decay with structural
  output deviations. Constraint violations are explicitly separated from
  format failures.}
\label{fig:pipeline}
\end{figure}

\CCB{} consists of three task domains, a deterministic generator that
produces ground-truth reasoning traces, a strict parser-based pipeline,
and a one-parameter decay model fit per (model, domain) cell. The
components are designed jointly: the failure taxonomy isolates the
events the decay model is meant to describe, the parser produces the
event counts the likelihood consumes, and the trace metric below
operates on the same parsed structure. Figure~\ref{fig:pipeline}
summarises the end-to-end evaluation flow, from dataset generation
through the strict parsing hierarchy that routes each trial into one
of the disjoint outcome categories used by the decay model.

\paragraph{Domains.}
The three domains share an experimental contract;  semantic content
held fixed, depth $N$ varied over $\{5,10,\dots,50\}$;  but stress
structurally distinct facets of long-horizon reasoning, together
approximating a minimal basis for the regimes most often encountered
in agentic settings: grounded spatial, abstract symbolic, and nested
relational. \emph{D1 Alien Grid} is a $3{\times}3$ grid subjected to
$N$ discrete transformations (\textsc{Rotate}, \textsc{Swap Corners},
\textsc{Shift}) under a grid-integrity constraint preventing entity
collision; it tests grounded spatial state-tracking, and its errors
compound deterministically because a misplaced entity at step $k$
invalidates every state after it. \emph{D2 Symbolic Pointer Tracking}
asks the model to maintain seven variables $A$--$G$ holding distinct
digits 0--9 under $N$ cyclic-shift and modular-arithmetic operations
subject to an assignment-uniqueness constraint; its dominant failure
mode is register corruption via illegal re-assignment, which accounts
for $69.5\%$ of D2 failures and reflects the difficulty of maintaining
disjoint symbolic mappings over long contexts. \emph{D3 Social Logic}
runs a diplomatic graph over ten agents under transitive-closure
rules: at each of $N$ update steps a new alliance or rivalry edge is
added, the closure is recomputed, and all implied relationships are
updated. The model is queried on the final pairwise state in a
Theory-of-Mind-style format (``what does agent $i$ believe about agent
$j$?''), and must maintain global pairwise consistency over $O(n^2)$
relationships per step. Its dominant failure mode is \emph{cascade collapse};  a
single misclassified edge propagates by transitivity to every
reachable node and is irrecoverable within the context window, which
is what makes D3 a structurally distinct difficulty regime rather
than a harder version of D1 or D2 (Section~\ref{sec:results}).
We evaluate each domain at $n{=}40$ independently seeded trials per
(model, depth) cell, for $2{,}000$ trials per domain and $6{,}000$ in
total. All inference is at $T{=}0$ via OpenRouter, on
\textit{claude-3.7-sonnet}, \textit{gemini-2.0-flash-001},
\textit{deepseek-chat}, \textit{gpt-4o-mini}, and
\textit{llama-3.3-70b-instruct}. Reasoning-specialised models
(o1/o3, DeepSeek-R1) were not accessible at submission time and are
addressed in Section~\ref{sec:discussion}.

\paragraph{Failure taxonomy and parser design.}
Every trial falls into one of six disjoint categories: \emph{Correct}
(final answer and every step exact), \emph{Reasoning} (parses cleanly
but diverges from the canonical trace at some step $\kstar{\leq}N$),
\emph{Constraint} (violates a structural task rule), \emph{Format}
(unparseable), \emph{Truncation} (output ends mid-stream), and \emph{API}
(network error, auto-retried). Only the first three carry information
about per-step retention; the rest are treated as missing data. The
parser is regex-based and deliberately so: AST-based parsing and
LLM-as-judge scoring were both considered and rejected to keep the
evaluation deterministic and reproducible, and to remove any dependency
on an external scoring model whose own reliability would have to be
defended. The parser prioritises precision over recall;  minor format
deviations are flagged as Format failures rather than silently
corrected, so the resulting $\pd$ estimates are mildly conservative
for models with idiosyncratic output styles. This is the bias direction
we want because it cannot inflate reported accuracy or $\pd$. A manual
audit of 150 edge-case outputs found no false positives in correctness
classification and a small ($\approx 2\%$) false-negative rate from
non-standard separators. The six-way classification is invariant to
reordering of unrelated key/value pairs and to whitespace differences,
so layout-only output changes do not affect the reported $\kstar$
distributions.

\paragraph{The TFBC metric.}
Correct final answers do not imply correct intermediate reasoning. We
define \emph{Trace First Branch Correct} (\textsc{TFBC}) to flag any
trial whose final answer matches the ground truth but whose trace first
diverges from the canonical reasoning at some step $\kstar{>}0$.
Algorithm~\ref{alg:tfbc} walks the trace from step 1 and returns the
first divergence; for incorrect parseable trials, $\kstar$
records the depth at which the model's working state first decohered.
As a concrete illustration, a typical $\kstar{=}2$ event on D1 at
$N{=}10$ has ground truth Step~2 $= [[7,4,1],[8,5,2],[9,6,3]]$ (a
$90^{\circ}$ clockwise rotation) and model output $[[3,2,1],[6,5,4],
[9,8,7]]$ (a horizontal flip); the trace diverges immediately at the
first transformation and the error propagates deterministically through
the rest of the trial, even when the final answer happens to coincide
by chance. Three human annotators independently labelled traces against
the automated extractor on subsets of each domain, with Cohen's
$\kappa$ of $0.977$ (D1, $n{=}50$), $0.978$ (D2, $n{=}50$), and $0.938$
(D3, $n{=}65$), all above the conventional defensibility threshold of
$0.80$. D3 admits multiple valid reasoning paths in principle; a
targeted audit of 20 randomly sampled D3 TFBC cases found no genuine
alternative-path correctness, supporting the lucky-guess interpretation.
Constraint violations and TFBC events are measured on different
conditioning sets (TFBC is defined only over parseable correct outputs;
constraint violations are a disjoint failure category) and so the high
D2 constraint-violation share and the per-model TFBC rates in
Table~\ref{tab:master_results} are not in tension.

\begin{algorithm}[h]
\caption{TFBC and $\kstar$ extractor.}
\label{alg:tfbc}
\begin{algorithmic}[1]
\STATE \textbf{Input:} model trace $T$, ground-truth trace $G$, depth $N$
\STATE $L \leftarrow \text{ParseSteps}(T)$;\ $\kstar \leftarrow -1$
\IF{$|L| < N$ \AND missing answer} \RETURN Format Error \ENDIF
\FOR{$i = 1$ \TO $N$}
  \IF{$L[i] \neq G[i]$} \STATE $\kstar \leftarrow i$;\ \textbf{break} \ENDIF
\ENDFOR
\STATE $\text{is\_TFBC} \leftarrow (\kstar \neq -1) \land (A_{\text{model}} = A_{\text{true}})$
\RETURN $\kstar$, $\text{is\_TFBC}$
\end{algorithmic}
\end{algorithm}

\paragraph{Geometric decay model.}
Let $\pd$ denote the probability that the model correctly computes
$S_k{\to}S_{k+1}$ given that $S_k$ was maintained. Under independence of
per-step errors (Assumption~\ref{ass:indep}),
\begin{equation}
P(\text{correct}\mid N)\;=\;\prod_{i=1}^N P(\text{step}_i\text{ correct})\;=\;\pd^N.
\label{eq:geometric}
\end{equation}
The expression $\pd^N$ is a discrete survival function with constant
per-step hazard $1{-}\pd$; the interpretation we want is that a single
free parameter per (model, domain) cell captures both the per-step
failure rate and the resulting horizon at which any chosen success
threshold is crossed. We fit $\pd$ per cell by maximising the binomial
log-likelihood $\sum_N\bigl[c_N\ln\pd^N+(n{-}c_N)\ln(1{-}\pd^N)\bigr]$
over $c_N$ correct trials out of $n{=}40$ at each depth, constrained to
$\pd{\in}[0.5,1.0]$; 95\% confidence intervals are from $2{,}000$
parametric bootstrap resamples. The lower bound keeps the estimator in
a meaningful regime: models driven to it (GPT-4o-mini and LLaMA on D3)
should be read as exhibiting no statistically identifiable
step-retention rather than as literal $50\%$ per-step.

\begin{assumption}[Independent per-step failures]
\label{ass:indep}
The probability of a state-transition error at step $k$ is independent
of whether an error occurred at step $k{-}1$.
\end{assumption}

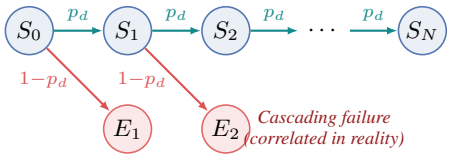
\begin{figure}[h]
\centering
\begin{tikzpicture}[
    every node/.style={font=\footnotesize},
    state/.style={circle,draw=accentnavy!70,fill=accentnavy!8,
        minimum size=18pt,inner sep=0pt,line width=0.6pt},
    fail/.style={circle,draw=deepseekc!70,fill=deepseekc!10,
        minimum size=18pt,inner sep=0pt,line width=0.6pt,font=\footnotesize\bfseries}]
\node[state] (s0) at (0,0)   {$S_0$};
\node[state] (s1) at (1.3,0) {$S_1$};
\node[state] (s2) at (2.6,0) {$S_2$};
\node       (sd) at (3.9,0) {$\cdots$};
\node[state] (sn) at (5.2,0) {$S_N$};
\node[fail]  (e1) at (1.3,-1.3) {$E_1$};
\node[fail]  (e2) at (2.6,-1.3) {$E_2$};
\draw[-{Latex[length=4pt]},thick,accentteal!85!black] (s0)--(s1) node[midway,above,font=\scriptsize]{$\pd$};
\draw[-{Latex[length=4pt]},thick,accentteal!85!black] (s1)--(s2) node[midway,above,font=\scriptsize]{$\pd$};
\draw[-{Latex[length=4pt]},thick,accentteal!85!black] (s2)--(sd) node[midway,above,font=\scriptsize]{$\pd$};
\draw[-{Latex[length=4pt]},thick,accentteal!85!black] (sd)--(sn) node[midway,above,font=\scriptsize]{$\pd$};
\draw[-{Latex[length=4pt]},thick,deepseekc!80] (s0)--(e1) node[midway,left,font=\scriptsize]{$1{-}\pd$};
\draw[-{Latex[length=4pt]},thick,deepseekc!80] (s1)--(e2) node[midway,left,font=\scriptsize]{$1{-}\pd$};
\node[font=\scriptsize\itshape,text=deepseekc!70!black,align=center]
   at (3.9,-1.3) {Cascading failure\\(correlated in reality)};
\end{tikzpicture}
\caption{\textbf{The state-retention process of Assumption~\ref{ass:indep}.}
  Under independence, a single step error transitions the system to
  an absorbing failure state with per-step probability $1{-}\pd$. In
  reality, errors at step $\kstar$ corrupt all $k{>}\kstar$
  (Remark~\ref{rem:corr}), so the true decay is faster than $\pd^N$
  predicts.}
\label{fig:markov}
\end{figure}

\paragraph{Alternative decay models.}
Assumption~\ref{ass:indep} is wrong in a known direction
(Figure~\ref{fig:markov}). Autoregressive transformers are not Markov:
an error at step $\kstar$ corrupts context for all $k{>}\kstar$, so
per-step errors are positively correlated and Equation~\ref{eq:geometric}
tends to \emph{overestimate} sustained accuracy. The TFBC phenomenon
- partial recovery from a corrupted trace to a correct final answer
- softens but does not eliminate this bias, and we treat $\pd^N$ as
an optimistic envelope
throughout. We considered two alternative forms before adopting the
geometric model: an accelerating $\pd^{N^\gamma}$ that would capture
attention fatigue, and a linear $1{-}\lambda N$ drift. The linear form
fails to capture the non-linear collapse observed at moderate $N$; the
accelerating form adds a free parameter without empirical motivation
in the $N{\leq}50$ regime. Across all three domains the one-parameter
geometric form fits without exception and explains $>90\%$ of accuracy
variance on D1 and D2 (a residual analysis confirms no systematic
structure in the depth-wise residuals). Formal AIC/BIC selection
between these formulations beyond $N{=}50$, where positional-embedding
saturation may shift the dominant failure mechanism, is left for
future work.

\paragraph{Horizon connection.}
The decay model also feeds the horizon-length analysis of
\citet{sinha2025illusion}: for minimum success threshold $s$, the
effective task horizon is $H_s{\approx}\ln(s)/\ln(\pd)$. For Claude on
D3 with $\pd{=}0.863$, $H_{0.5}{\approx}4.7$ steps, which matches both
the depth at which observed accuracy crosses $50\%$ and the mean
$\kstar{=}4.30$ on incorrect D3 trials. The empirical $\pd$ thus has a
deployment-facing reading: the depth at which a model's expected
accuracy on a task family falls below any chosen threshold.

\section{Results}
\label{sec:results}

Table~\ref{tab:master_results} summarises all 6{,}000 evaluations and
Figure~\ref{fig:hero} shows the depth-resolved accuracy. The data
support three claims: per-step retention decays geometrically across
all five models and three domains; the D3 ceiling is qualitatively
different from the D1/D2 ceiling and is not movable by prompt-level
intervention; and the trace-divergence step $\kstar$ ranks models
more faithfully within domain than parameter count or aggregate
accuracy.

\begin{table*}[t]
\centering
\caption{\textbf{\CCB{} master results.} Aggregate accuracy, the
  geometric step-retention MLE $\pd$, and the trace-level lucky-guess
  rate \textsc{TFBC}, per (model, domain) cell. Best per domain in
  \textbf{bold}; per-depth tables with 95\% bootstrap CIs are in
  Appendix~\ref{app:detailed_results}. $^{\dagger}$ At the optimiser lower
  bound; no statistically identifiable step-retention.}
\label{tab:master_results}
\small
\setlength{\tabcolsep}{4pt}
\begin{tabular}{l | ccc | ccc | ccc | c}
\toprule
\multirow{2}{*}{\textbf{Model}}
  & \multicolumn{3}{c|}{\textbf{D1 Spatial}}
  & \multicolumn{3}{c|}{\textbf{D2 Symbolic}}
  & \multicolumn{3}{c|}{\textbf{D3 Social Logic}}
  & \multirow{2}{*}{\textbf{Avg}}\\
\cmidrule(lr){2-4}\cmidrule(lr){5-7}\cmidrule(lr){8-10}
 & Acc & $\pd$ & TFBC
 & Acc & $\pd$ & TFBC
 & Acc & $\pd$ & TFBC & \\
\midrule
\multicolumn{11}{l}{\textit{Frontier / closed-weight}}\\
Claude 3.7    & 18.2\% & 0.929 & 21\%
              & \textbf{71.2\%} & \textbf{0.987} & 8\%
              & \textbf{6.2\%}  & \textbf{0.863} & 56\%
              & \textbf{31.9\%} \\
Gemini 2.0F   & \textbf{22.0\%} & \textbf{0.930} & 1\%
              & 30.0\% & 0.950 & 29\%
              & 3.2\%  & 0.736 & 62\%
              & 18.4\% \\
GPT-4o-mini   & 1.5\%  & 0.688 & 17\%
              & 1.2\%  & 0.634 & 20\%
              & 0.2\%  & 0.500$^\dagger$ & --
              & 1.0\%  \\
\midrule
\multicolumn{11}{l}{\textit{Open-weight}}\\
DeepSeek      & 19.8\% & 0.924 & 6\%
              & 31.3\% & 0.955 & 12\%
              & 2.0\%  & 0.684 & 13\%
              & 17.7\% \\
LLaMA-3.3     & 1.3\%  & 0.634 & 0\%
              & 4.3\%  & 0.767 & 18\%
              & 0.0\%  & 0.500$^\dagger$ & --
              & 1.9\%  \\
\bottomrule
\end{tabular}
\end{table*}

\begin{figure*}[t]
\centering
\newcommand{\hmrow}[3]{%
  \node[anchor=east,font=\scriptsize] at (-0.08, 2.1 - 0.36*#1) {#2};%
  \foreach \v [count=\d from 0] in {#3} {%
    \pgfmathsetmacro{\shade}{min(100, max(\v, \v*0.9))}%
    \fill[accentnavy!\shade] (0.42*\d, 2.1 - 0.36*#1 - 0.16)
          rectangle ++(0.40,0.32);%
    \ifdim \v pt > 0pt
      \pgfmathtruncatemacro{\vint}{round(\v)}%
      \ifdim \v pt > 45pt
        \node[font=\tiny\bfseries,text=white] at (0.42*\d + 0.20, 2.1 - 0.36*#1) {\vint};%
      \else
        \node[font=\tiny,text=black] at (0.42*\d + 0.20, 2.1 - 0.36*#1) {\vint};%
      \fi
    \fi
  }%
}
\newcommand{\hmpanel}[1]{%
  \node[anchor=south,font=\small\bfseries] at (2.1,2.55) {#1};%
  \foreach \d [count=\i from 0] in {5,10,15,20,25,30,35,40,45,50} {%
    \node[anchor=north,font=\tiny] at (0.42*\i + 0.20, -0.05) {\d};%
  }%
  \node[anchor=north,font=\scriptsize] at (2.0,-0.42) {Depth $N$};%
}
\begin{tikzpicture}[font=\footnotesize]
\begin{scope}[shift={(0,0)}]
  \hmpanel{D1 Alien Grid}
  \hmrow{0}{Claude}{57.5,25,15,7.5,20,10,15,7.5,15,10}
  \hmrow{1}{Gemini}{65,55,35,10,17.5,10,17.5,7.5,2.5,0}
  \hmrow{2}{DeepSeek}{70,42.5,22.5,10,22.5,7.5,10,10,2.5,0}
  \hmrow{3}{GPT-4o-m}{10,2.5,2.5,0,0,0,0,0,0,0}
  \hmrow{4}{LLaMA}{12.5,0,0,0,0,0,0,0,0,0}
\end{scope}
\begin{scope}[shift={(5.6,0)}]
  \hmpanel{D2 Symbolic Pointers}
  \hmrow{0}{Claude}{97.5,90,85,67.5,80,57.5,70,57.5,52.5,55}
  \hmrow{1}{Gemini}{55,52.5,50,42.5,25,20,22.5,5,10,17.5}
  \hmrow{2}{DeepSeek}{37.5,30,45,52.5,30,35,30,17.5,22.5,12.5}
  \hmrow{3}{GPT-4o-m}{12.5,0,0,0,0,0,0,0,0,0}
  \hmrow{4}{LLaMA}{40,2.5,0,0,0,0,0,0,0,0}
\end{scope}
\begin{scope}[shift={(11.2,0)}]
  \hmpanel{D3 Social Logic}
  \hmrow{0}{Claude}{35,5,0,0,5,7.5,2.5,2.5,0,5}
  \hmrow{1}{Gemini}{32.5,0,0,0,0,0,0,0,0,0}
  \hmrow{2}{DeepSeek}{20,0,0,0,0,0,0,0,0,0}
  \hmrow{3}{GPT-4o-m}{2.5,0,0,0,0,0,0,0,0,0}
  \hmrow{4}{LLaMA}{0,0,0,0,0,0,0,0,0,0}
\end{scope}
\begin{scope}[shift={(16.7,0.3)}]
  \foreach \v in {0,10,...,100} {
    \fill[accentnavy!\v] (0,\v*0.018) rectangle ++(0.22,0.020);
  }
  \draw[black!60] (0,0) rectangle (0.22,1.98);
  \node[anchor=west,font=\tiny] at (0.24,0)    {0\%};
  \node[anchor=west,font=\tiny] at (0.24,0.99) {50\%};
  \node[anchor=west,font=\tiny] at (0.24,1.98) {100\%};
  \node[anchor=south,font=\scriptsize,rotate=90] at (-0.18,1.0) {Accuracy};
\end{scope}
\end{tikzpicture}
\caption{\textbf{Per-depth accuracy across models and domains.}
  Rows are models, columns are depth $N{\in}\{5,\dots,50\}$, cells
  show observed accuracy (\%), darker = higher. The qualitative
  difference between domains is immediate: D1 and D2 retain
  non-trivial gradients at large $N$ for frontier models, while D3 is
  essentially a single column of non-zero values at $N{=}5$. The
  view complements Table~\ref{tab:master_results} (which aggregates
  across depths) and Figure~\ref{fig:hero} (which shows decay
  curves).}
\label{fig:heatmap}
\end{figure*}
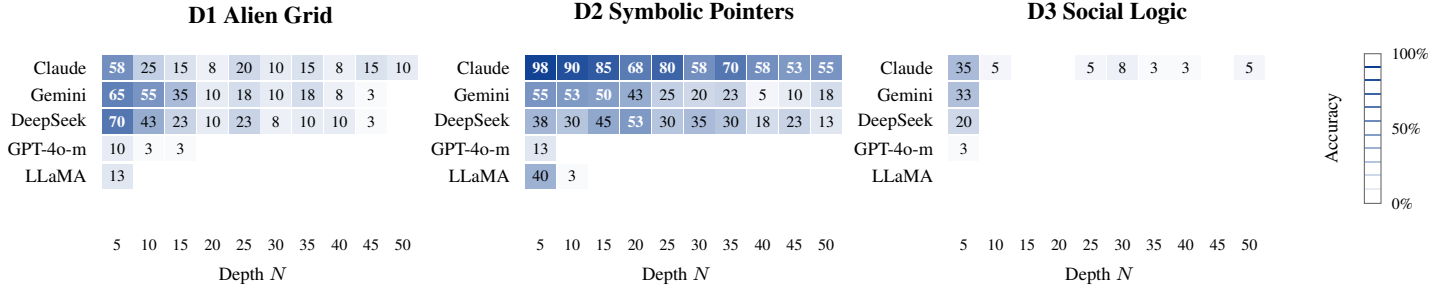

\paragraph{D1 and D2: per-step retention.}
The decay curves separate the five models cleanly on D1 and D2. The
three strongest models cluster around $\pd \in [0.92, 0.93]$ on D1
while GPT-4o-mini ($\pd{=}0.688$) and LLaMA-3.3 ($\pd{=}0.634$) collapse
by $N{=}15$. D1 format adherence is essentially perfect (0--2\% format
failures), so the gap between tiers is in per-step retention rather than
output structure. D2 sharpens the picture: Claude reaches
$\pd{=}0.987$ and $71.2\%$ aggregate accuracy, more than double Gemini
($30.0\%$) or DeepSeek ($31.3\%$), and retains $55\%$ accuracy at
$N{=}50$. Claude's D2 decay curve is visibly shallower than the
non-frontier models';  a $\pd{=}0.987$ vs.\ $\pd{=}0.634$ gap
corresponds to roughly a $28{\times}$ lower per-step error rate, which
is exactly the regime in which long-horizon agentic use becomes
plausible. Of all D2 failures, $69.5\%$ are constraint violations -
illegal re-assignments of variables already bound. A canonical instance
is a trace that produces \texttt{Step 4: var\_A=10}, \texttt{Step 5:
var\_B=var\_A} (so $\texttt{var\_B}$ should hold 10 for the rest of
the trial), then writes \texttt{Step 6: var\_A=15}, silently violating
uniqueness on $\texttt{var\_A}$. The model is performing arithmetic
correctly at each step but failing to keep its bindings disjoint over
the longer horizon, which is exactly the failure mode the
assignment-uniqueness constraint was designed to expose. The
occasional non-monotonic accuracy upticks at
$N{=}25$--$35$ on D1 and D2 are within the per-cell Clopper-Pearson
half-CI of $\pm 12$--$16\%$ at $n{=}40$ and do not reflect generator
artefacts: operation distributions are stationary across $N$ by
construction, and the monotone $\pd^N$ fit explains $>90\%$ of variance
on these domains.

\paragraph{D3 (transitive closure).}
D3 is qualitatively different. Across all five models and ten depth
levels, $1{,}953$ of $2{,}000$ attempts fail. Only Claude achieves any
sustained accuracy ($6.2\%$ overall, concentrated at $N{=}5$); every
other model collapses to near-zero past depth 5 regardless of capability
tier. The mean step at which reasoning first diverges, $\kstar$, is
uniformly low on D3 (2.88--4.30) across \emph{all} models, including
Claude;  which carries $\kstar{=}17.67$ on D2 yet diverges after
only 4.30 steps on D3 (Table~\ref{tab:kstar}). The uniformity of
$\kstar$ across models with markedly different capability on other
domains is direct evidence that D3 poses a
qualitatively distinct difficulty.

\paragraph{Why does D3 collapse early?}
This is consistent with the computational structure of D3, which
differs from D1 and D2 along three dimensions. First, transitive
closure is not decomposable into independent per-step updates: a
single misclassified edge at step $k$ propagates by transitivity to
every $k$-hop reachable node, so unlike D1 (where errors compound
locally) or D2 (where bindings can in principle be reread), the D3
state offers no per-step recovery path. Second, the task requires
maintaining global pairwise consistency over $O(n^2)$ relationships
per update, which competes for representational capacity with the
linear context attention exposes. Third;  and this is the operative
mechanical claim;  standard attention flattens sequence hierarchy
into linear context, providing no mechanism for the kind of recursive
stack management that transitive closure demands; as $N$ grows the
representations of distinct agents' relationship sets mix in the
attention pattern rather than remaining cleanly partitioned, and the
divergence step $\kstar{\in}[2.88, 4.30]$ across models of very
different capability is consistent with the prediction that this
ceiling reflects an architectural rather than a capacity bottleneck.
We offer this as an account of the collapse, not a proof of
impossibility: tool-augmented systems with explicit graph state, or
process-supervised models with step-level reward, could plausibly
move the ceiling \citep{yang2026recursive, takashiro2026thinking}.
\CCB{} probes reasoning without such scaffolding; quantifying whether
the ceiling moves when scaffolding is added is one of the uses the
benchmark is built to serve. As a concrete illustration of the
failure mode, a representative D3 cascade trace from DeepSeek at
$N{=}10$ processes the first update correctly but omits a transitive
closure propagation at step 2 (failing to mark a pair as allied that
inherits the relation through a newly added edge); every subsequent
pair classification inherits this corruption, and the model has no
mechanism for retroactively correcting the state from within the
context window.

\begin{table*}[t]
\centering
\caption{\textbf{Per-domain results with format-failure rates and
  $\pd$ bootstrap 95\% CI widths.} The three domains are broken out
  separately so the per-domain spread, the uncertainty on $\pd$, and
  the format-failure share are all directly inspectable. CI widths
  ($\Delta\pd$) are the full 95\% interval widths from 2{,}000
  parametric bootstrap resamples; intervals straddle the point
  estimate symmetrically except where the optimiser lower bound
  $\pd{=}0.5$ truncates them. $\kappa$ values are inter-annotator
  agreement on the $\kstar$ extractor. Best per domain in
  \textbf{bold}. $^{\dagger}\pd$ at the optimiser lower bound (no
  statistically identifiable step-retention).}
\label{tab:per_domain}
\footnotesize
\setlength{\tabcolsep}{3pt}
\begin{tabular}{@{}lcccc@{\quad}lcccc@{\quad}lcccc@{}}
\toprule
\multicolumn{5}{c}{\textbf{D1 Alien Grid} ($\kappa{=}0.977$)}
& \multicolumn{5}{c}{\textbf{D2 Symbolic Pointers} ($\kappa{=}0.978$)}
& \multicolumn{5}{c}{\textbf{D3 Social Logic} ($\kappa{=}0.938$)} \\
\cmidrule(lr){1-5}\cmidrule(lr){6-10}\cmidrule(lr){11-15}
Model & Acc & $\pd$ & $\Delta\pd$ & TFBC
& Model & Acc & $\pd$ & $\Delta\pd$ & TFBC
& Model & Acc & $\pd$ & $\Delta\pd$ & TFBC \\
\midrule
Gemini   & \textbf{22.0\%} & 0.930 & 0.018 & 1\%
& Claude   & \textbf{71.2\%} & \textbf{0.987} & 0.004 & 8\%
& Claude   & \textbf{6.2\%}  & \textbf{0.863} & 0.046 & 56\% \\
DeepSeek & 19.8\%          & 0.924 & 0.021 & 6\%
& DeepSeek & 31.3\%          & 0.955 & 0.012 & 12\%
& Gemini   & 3.2\%           & 0.736 & 0.118 & 62\% \\
Claude   & 18.2\%          & 0.929 & 0.019 & 21\%
& Gemini   & 30.0\%          & 0.950 & 0.012 & 29\%
& DeepSeek & 2.0\%           & 0.684 & 0.167 & 13\% \\
GPT-4o-m & 1.5\%           & 0.688 & 0.165 & 17\%
& LLaMA    & 4.3\%           & 0.767 & 0.099 & 18\%
& GPT-4o-m & 0.2\%           & 0.500$^\dagger$ & -- & --  \\
LLaMA    & 1.3\%           & 0.634 & 0.205 & 0\%
& GPT-4o-m & 1.2\%           & 0.634 & 0.208 & 20\%
& LLaMA    & 0.0\%           & 0.500$^\dagger$ & -- & --  \\
\bottomrule
\end{tabular}
\end{table*}

\begin{table}[h]
\caption{Mean divergence step $\kstar$ for incorrect trials. Within
  domain, $\kstar$ tracks accuracy; across models, parameter count
  does not;  LLaMA-3.3 (70B) sits below Claude on every domain.}
\label{tab:kstar}
\centering\small
\setlength{\tabcolsep}{6pt}
\begin{tabular}{lccc}
\toprule
Model & D1 $\kstar$ & D2 $\kstar$ & D3 $\kstar$ \\
\midrule
Claude 3.7   & 8.45 & \textbf{17.67} & \textbf{4.30} \\
Gemini 2.0F  & \textbf{9.22} & 11.85 & 3.36 \\
DeepSeek     & 8.10 &  9.92 & 3.45 \\
LLaMA-3.3    & 3.45 &  3.91 & 3.01 \\
GPT-4o-mini  & 3.21 &  3.30 & 2.88 \\
\bottomrule
\end{tabular}
\end{table}

\paragraph{Trace-faithful vs.\ lucky-guess correctness.}
A second axis of dissociation appears when we look at correct outputs
rather than incorrect ones. A trace-level view reveals two qualitatively
different populations among the answers an output-only evaluator would
treat identically. \emph{Trace-faithful correctness} ($\kstar{=}{-}1$,
TFBC false) is correctness with intact intermediate reasoning;
\emph{lucky-guess correctness} (TFBC true) is correctness despite
demonstrably divergent reasoning. The dominant population shifts with
domain. On D2, where Claude operates near $\pd{=}0.99$, only $8\%$ of
its correct outputs are TFBC and the rest are genuinely faithful
(262 of 285 traces match the canonical reasoning); D2 is dominated by
trace-faithful correctness. On D3 the picture inverts: $56\%$--$62\%$
of correct outputs from Claude and Gemini are TFBC, and the targeted
audit (Section~\ref{sec:framework}) found no genuine alternative-path
correctness in this group, so D3 correctness is dominated by
lucky-guess events. Aggregated across the benchmark, $14.5\%$ of
correct outputs are TFBC, so output-only evaluation overstates
reasoning quality \emph{and} does so differentially;  with the
overstatement concentrated on precisely the harder domains where it
most matters. A reader of the aggregate scores would conclude that
Claude's $6.2\%$ D3 accuracy reflects real partial competence; the
trace-level evidence says that conclusion would be wrong for most of
that $6.2\%$.

\begin{figure}[h]
\centering
\begin{tikzpicture}
\begin{axis}[
    xbar stacked, width=\linewidth, height=4.6cm,
    xlabel={\% of D2 failures},
    symbolic y coords={LLaMA, GPT-4o-m, DeepSeek, Gemini, Claude},
    ytick=data,
    tick label style={font=\scriptsize},
    label style={font=\scriptsize},
    legend style={font=\scriptsize, at={(0.5,-0.32)},
      anchor=north, legend columns=-1, draw=none},
    xmin=0, xmax=100, bar width=10pt,
    axis line style={draw=black!60}
]
\addplot[fill=deepseekc!50,draw=deepseekc!70] coordinates {
  (52.7,LLaMA)(61.9,GPT-4o-m)(29.2,DeepSeek)(7.7,Gemini)(6.5,Claude)};
\addplot[fill=accentamber!50,draw=accentamber!80] coordinates {
  (45.5,LLaMA)(37.1,GPT-4o-m)(38.2,DeepSeek)(46.3,Gemini)(28.3,Claude)};
\addplot[fill=geminic!50,draw=geminic!70] coordinates {
  (1.9,LLaMA)(1.0,GPT-4o-m)(32.6,DeepSeek)(46.0,Gemini)(65.2,Claude)};
\legend{Early ($\kstar{\leq}3$), Mid, Late ($\kstar{>}10$)}
\end{axis}
\end{tikzpicture}
\caption{Divergence-step distribution on D2 by model. LLaMA fails at
  the first symbolic transition; Claude maintains accuracy across
  $\sim$17 steps before failing on global consistency.}
\label{fig:tfbc_hist}
\end{figure}
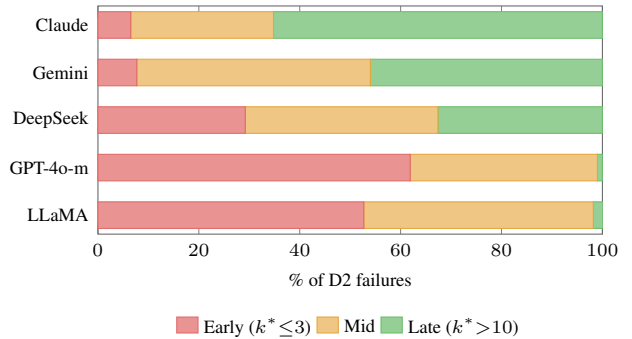

\paragraph{$\kstar$ as a coherence-depth statistic.}
The distribution of $\kstar$, not just its mean, carries information.
Figure~\ref{fig:tfbc_hist} shows D2 stratified by model: LLaMA fails
early at the first symbolic transition, while Claude maintains
accuracy across $\sim$17 steps before failing on global consistency.
Early-heavy and late-heavy failure modes are qualitatively different
and would call for different mitigations even at matched aggregate
accuracy.

Taken together, $\kstar$ functions as a \emph{working-memory-coherence
depth statistic}: an operational measurement of how many sequential
state updates a model can compose before its working representation
decoheres. This framing explains the empirical regularity that
parameter count is a poor cross-model predictor.
LLaMA-3.3-70B has $70$B parameters;  more than most of the
closed-weight comparators in this study;  yet $\kstar_\text{LLaMA}{<}\kstar_\text{Claude}$ on every
domain ($3.45$ vs.\ $8.45$ on D1, $3.91$ vs.\ $17.67$ on D2, $3.01$
vs.\ $4.30$ on D3). Within a domain, $\kstar$ tracks accuracy because
the depth at which a trace first decoheres mechanically lower-bounds
the accuracy achievable beyond that depth; across models, the
$\kstar$ ranking tracks the kind of reasoning capability that drives
long-horizon agentic performance more faithfully than scale. We
therefore propose $\kstar$ as a complement, not a replacement, to
aggregate accuracy: a single (model, task-family) cell reports both
$\pd$ (the per-step retention) and $\kstar$ (the typical
coherence depth) without requiring per-depth evaluation at use time.

\paragraph{Verbosity ablation.}
A natural objection is that the D3 collapse reflects prompt phrasing
rather than a genuine reasoning limit, since prompt sensitivity in LLM
benchmarks is well documented and the cliff appears uniformly at
$N{=}5$. We tested this with a paired ablation at $N{=}15$ on Claude,
the only model with any D3 signal. The Standard condition lets the
model infer state naturally; the Verbose condition prepends a
\textsc{critical} instruction forcing it to restate the entire agent
belief array after every operation. Both conditions yielded $0.0\%$
accuracy across all $n{=}20$ paired instances, with the McNemar
contingency in Table~\ref{tab:mcnemar}: zero discordant pairs and
McNemar $p{=}1.000$. Token usage tells the same story;  the Verbose
condition spent $\sim$1{,}362 tokens to the Standard's $\sim$1{,}282
(a $6\%$ overhead). The model acknowledged the instruction but spent
the extra tokens restating beliefs it could not compute correctly,
not on doing the computation differently. With zero discordant pairs
the McNemar test confirms equipotence rather than distinguishing
architecturally caused failure from coincidentally uniform failure,
so we treat the result as suggestive rather than dispositive.

\begin{table}[h]
\centering
\caption{D3 verbosity ablation contingency, $N{=}15$, Claude,
  $n{=}20$ paired. McNemar $p{=}1.000$.}
\label{tab:mcnemar}
\small
\setlength{\tabcolsep}{8pt}
\begin{tabular}{lcc}
\toprule
 & Verbose Correct & Verbose Wrong \\
\midrule
Standard Correct & 0 & 0 \\
Standard Wrong   & 0 & 20 \\
\bottomrule
\end{tabular}
\end{table}

\begin{figure}[h]
\centering
\begin{tikzpicture}
\begin{axis}[
    ybar, width=\linewidth, height=3.4cm,
    enlargelimits=0.15,
    ylabel={Accuracy (\%)},
    symbolic x coords={Standard, Verbose, Var A, Var B, Var C},
    xtick=data,
    nodes near coords, nodes near coords align={vertical},
    tick label style={font=\scriptsize},
    label style={font=\scriptsize},
    ymin=0, ymax=40, bar width=15pt,
    axis line style={draw=black!60}
]
\addplot[fill=claudec!50,draw=claudec!70,thick]
  coordinates {(Standard,0.0)(Verbose,0.0)(Var A,0.0)
               (Var B,20.0)(Var C,0.0)};
\end{axis}
\end{tikzpicture}
\caption{\textbf{D3 prompt-sensitivity at $N{=}15$} (Claude 3.7,
  $n{=}20$ per condition). Standard/Verbose yield identical $0\%$
  (McNemar $p{=}1.000$); only \textsc{Var B} (positional slot
  formatting) reaches non-zero accuracy, and at $20.0\%$ this still
  falls far short of practical utility on long-horizon tasks.}
\label{fig:ablation}
\end{figure}
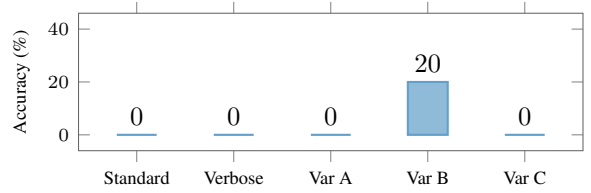

\paragraph{Prompt-structure ablation.}
To check whether \emph{any} prompt structure shifts the ceiling we ran
three further variants at the same $N{=}15$: \textsc{Var A} imposes a
strict output schema, \textsc{Var B} adds positional formatting
($\texttt{P\#\#}$ slot markers identifying each agent position), and
\textsc{Var C} adds an explicit logical mapping of the alliance/rivalry
update rule (Figure~\ref{fig:ablation}). Variants A and C remain at
$0.0\%$. Variant B reaches $20.0\%$;  a 20-point absolute spread that
demonstrates D3 is prompt-sensitive in principle, but still falls far
below practical utility. We do not yet have a mechanistic account of
why slot-based positional formatting partially succeeds where the other
variants do not, and whether the benefit extends to $N{>}15$ or to
other models is reserved for the next iteration. The qualitative
takeaway is the same as the Standard/Verbose result: prompt-level
interventions can move the D3 ceiling by a few percentage points but
do not change the underlying regime.

\paragraph{Summary of regimes.}
Pulled together, the three domains identify three structurally
distinct failure regimes that aggregate accuracy collapses into a
single number: per-step retention bottleneck on D1 (where the
frontier-vs.-non-frontier gap lies in $\pd$, not formatting); a
constraint-management bottleneck on D2 (where $69.5\%$ of failures
are illegal re-assignments and the dominant error is state-keeping
rather than arithmetic); and a structurally distinct cascade collapse
on D3 (where $\kstar$ is essentially uniform across capability tiers
and accuracy is dominated by lucky-guess events). No model dominates
all three: Gemini leads on D1 ($22.0\%$), Claude leads on D2
($71.2\%$) and D3 ($6.2\%$), and the within-tier gaps shrink sharply
as the load shifts from per-step retention to structural
consistency;  Claude's $\sim$53\,pp lead on D2 collapses to a
$\sim$3\,pp lead on D3, evidence that the D3 ceiling is not a simple
translation of the D2 ranking. The deployment-facing implication is
direct: a model with $\pd{<}0.93$ should not be relied on for tasks
requiring more than $\sim$20 sequential steps, with the optimism
caveat of Remark~\ref{rem:corr} in mind.

\section{Discussion}
\label{sec:discussion}

\paragraph{Three failure regimes, three mitigations.}
The combined empirical and analytical picture supports a
\emph{state-drift} view of autoregressive failure under depth scaling:
each of the three CCB domains stresses a different facet of the same
underlying step-retention bottleneck, and the three resulting failure
regimes call for qualitatively different mitigations. Output-format
retraining is the natural target for D1 (where the gap is in $\pd$
and format-failure rates are already at $0$--$2\%$);
constraint-reminder prompting or scratchpad-augmented decoding is the
natural target for D2 (where the dominant failure is illegal
re-assignment under a uniqueness rule); and architectural or
training-level intervention;  process supervision, recursive
scaffolding, or explicit graph state;  is the indicated direction
for D3, since the verbosity ablation shows no movement from
prompt-level changes alone. Because $\pd$ is a single interpretable
number per (model, domain) cell, it can be reported alongside
aggregate accuracy on deployment dashboards without re-running
depth-stratified evaluation at use time.

\paragraph{The D1 ceiling is not a formatting artefact.}
A skeptical reading of these results turns naturally to the D1
result. Format adherence is uniformly high across all five models
($0$--$2\%$ format failures), which rules out the obvious confound
that the frontier-vs.-non-frontier gap is an artefact of structured
output. Frontier models maintain $\pd \in [0.924, 0.930]$;
GPT-4o-mini ($\pd{=}0.688$) and LLaMA ($\pd{=}0.634$) collapse on
per-step retention, not on formatting. The implication is that the
D1 ceiling is not closeable by output-format prompting alone, and
the right diagnostic target for closing it is per-step retention
rather than output adherence. Whether targeted format-aware
fine-tuning shifts $\pd$ on D1 is a clean empirical question we
leave open.

\paragraph{Missing reasoning-specialised baselines.}
The principal limitation of this work is the absence of
reasoning-specialised baselines. Process-level supervision
\citep{cobbe2021training} is the most plausible single intervention
for shifting $\pd$ beyond the frontier tier on D2 or for reducing
the D3 cascade; if process-supervised models achieve $\kstar{>}10$
on D3 it would substantially qualify the architectural reading of
our findings. We commit to evaluating o1/o3, DeepSeek-R1, recursive
scaffold models \citep{yang2026recursive}, and fast-slow recurrent
mechanisms \citep{takashiro2026thinking} alongside tool-augmented
systems with explicit graph state in the next iteration. This is the
single most informative follow-up experiment available, and the one
most likely to sharpen or falsify the structural reading of D3.

\begin{remark}[The bound is optimistic]
\label{rem:corr}
An error at step $\kstar$ corrupts context for all $k{>}\kstar$, so
per-step errors are positively correlated and the true decay is
faster than $\pd^N$ predicts. The TFBC phenomenon;  partial
recovery from a corrupted context to a correct final answer -
softens but does not eliminate this effect.
\end{remark}

\paragraph{Limitations.}
The remaining limitations are quantitative and we have tried to bound
them rather than eliminate them. The independence assumption behind
$\pd^N$ is wrong in a known direction
(Assumption~\ref{ass:indep}, Remark~\ref{rem:corr}); real
autoregressive decoding produces positively correlated errors, so
the model's predictions should be read as an optimistic envelope on
sustained accuracy rather than a guarantee. The synthetic generators
produce structurally controlled tasks but their generalisation to
naturalistic agentic settings requires further study. The strict
regex parser may under-count valid traces with idiosyncratic
formatting (we measured $\approx 2\%$ false-negative rate in the
150-case audit). All evaluations were performed at $T{=}0$ via
OpenRouter; provider-specific optimisations may shift profiles
slightly. The D3 prompt ablation is preliminary (one model, $n{=}20$
paired) and a broader sweep over prompt structures is needed to
distinguish architectural from coincidental failure cleanly. Formal
AIC/BIC selection between geometric, accelerating
$\pd^{N^\gamma}$, and linear $1{-}\lambda N$ decay forms at
$N{>}50$;  where positional-embedding saturation may shift the
dominant failure mechanism;  remains future work. Finally, TFBC
assumes a single canonical reasoning trace per trial; alternative
valid traces would be misclassified, though the 20-case D3 audit
found no such cases. The $150$-case parser audit and the
$\kappa{\geq}0.938$ inter-annotator agreement support that the
reported numbers are not parser artefacts.

\section{Conclusion}
\label{sec:conclusion}

The Complexity Ceiling Benchmark isolates reasoning depth as a
controlled experimental variable and reveals three structurally
distinct failure regimes that aggregate accuracy collapses into a
single number: per-step retention collapse on grounded spatial
reasoning, constraint-management collapse on abstract symbolic
reasoning, and a transitive-closure cascade on relational reasoning
that persists uniformly across all five evaluated models regardless
of parameter count. The trace-level TFBC analysis shows that
$14.5\%$ of correct outputs across the benchmark are reached via
demonstrably divergent intermediate reasoning, so trace-faithful and
lucky-guess correctness must be distinguished for benchmark scores
to remain meaningful on long-horizon tasks. The mean trace-divergence
step $\kstar$ predicts within-domain accuracy more faithfully than
parameter count, supporting a state-drift rather than a capacity-limit
account of autoregressive failure under depth scaling. The most urgent
open question is whether process-supervised or recursive architectures
dissolve the D3 ceiling;  if they do, the structural reading of
these findings requires substantial revision; if they do not, \CCB{}
provides a principled diagnostic for the next generation of
memory-augmented systems.

\bibliographystyle{plainnat}
\bibliography{ccb_refs}

\clearpage
\appendix

\section{Detailed Results}
\label{app:detailed_results}

This appendix supplements the per-cell summary in
Table~\ref{tab:master_results} and the per-depth heatmap in
Figure~\ref{fig:heatmap} with full numeric values:
Tables~\ref{tab:d1}--\ref{tab:d3} summarise across depths with 95\%
bootstrap CIs, and Tables~\ref{app:d1_full}--\ref{app:d3_full} give
the underlying depth-wise accuracies with Clopper-Pearson half-CIs.

\begin{table}[h]
\caption{D1 Alien Grid results ($n{=}400$/model, LLaMA $n{=}395$).
  $\pd$ 95\% CI in brackets. Human $\kappa{=}0.977$, $n{=}50$.}
\label{tab:d1}
\centering\small\setlength{\tabcolsep}{4pt}
\begin{tabular}{lcccc}
\toprule
Model & Acc. & $\pd$ [95\% CI] & TFBC & Fmt\% \\
\midrule
Gemini 2.0F & \textbf{22.0\%} & 0.930 [0.920, 0.938] & 1\%  & 0.0\% \\
DeepSeek    & 19.8\% & 0.924 [0.912, 0.933] & 6\%  & 0.0\% \\
Claude 3.7  & 18.2\% & 0.929 [0.918, 0.937] & 21\% & 0.0\% \\
GPT-4o-mini & 1.5\% & 0.688 [0.581, 0.746] & 17\% & 0.0\% \\
LLaMA-3.3   & 1.3\% & 0.634 [0.500, 0.705] & 0\%  & 2.0\% \\
\bottomrule
\end{tabular}
\end{table}

\begin{table}[h]
\caption{D2 Symbolic Pointer Tracking ($n{=}400$/model, DeepSeek
  $n{=}399$, LLaMA $n{=}392$). Human $\kappa{=}0.978$, $n{=}50$.}
\label{tab:d2}
\centering\small\setlength{\tabcolsep}{4pt}
\begin{tabular}{lcccc}
\toprule
Model & Acc. & $\pd$ [95\% CI] & TFBC & Fmt\% \\
\midrule
Claude 3.7  & \textbf{71.2\%} & 0.987 [0.985, 0.989] & 8\%  & 0.0\% \\
DeepSeek    & 31.3\% & 0.955 [0.948, 0.960] & 12\% & 5.5\% \\
Gemini 2.0F & 30.0\% & 0.950 [0.943, 0.955] & 29\% & 4.2\% \\
LLaMA-3.3   & 4.3\%  & 0.767 [0.705, 0.804] & 18\% & 1.0\% \\
GPT-4o-mini & 1.2\%  & 0.634 [0.500, 0.708] & 20\% & 1.2\% \\
\bottomrule
\end{tabular}
\end{table}

\begin{table}[h]
\caption{D3 Social Logic ($n{=}400$/model). Human $\kappa{=}0.938$,
  $n{=}65$. McNemar verbosity ablation $p{=}1.0$. $\pd$ at $0.500$
  is at the optimiser lower bound.}
\label{tab:d3}
\centering\small\setlength{\tabcolsep}{4pt}
\begin{tabular}{lcccc}
\toprule
Model & Acc. & $\pd$ [95\% CI] & TFBC & Fmt\% \\
\midrule
Claude 3.7  & \textbf{6.2\%} & 0.863 [0.836, 0.882] & 56\% & 0.2\% \\
Gemini 2.0F & 3.2\%  & 0.736 [0.663, 0.781] & 62\% & 0.2\% \\
DeepSeek    & 2.0\%  & 0.684 [0.576, 0.743] & 13\% & 4.8\% \\
GPT-4o-mini & 0.2\%  & 0.500$^\dagger$ [0.500, 0.611] & $*$ & 0.0\% \\
LLaMA-3.3   & 0.0\%  & 0.500$^\dagger$ [0.500, 0.611] & --  & 0.0\% \\
\bottomrule
\multicolumn{5}{l}{\footnotesize $*$ Single correct instance; TFBC
  unreliable.}\\
\end{tabular}
\end{table}

\begin{table*}[!t]
\centering
\caption{D1 Alien Grid: Acc\% $\pm$ Clopper-Pearson half-CI by depth.
  AvgTok = mean response tokens.}
\label{app:d1_full}
\small
\begin{tabular}{lcccccc}
\toprule
\textbf{N} & Claude & Gemini & DeepSeek & GPT-4o-m & LLaMA & AvgTok \\
\midrule
 5 & 57.5$\pm$16.0 & 65.0$\pm$15.5 & 70.0$\pm$15.0 & 10.0$\pm$10.4 & 12.5$\pm$11.3 & 183 \\
10 & 25.0$\pm$14.3 & 55.0$\pm$16.1 & 42.5$\pm$16.0 &  2.5$\pm$ 6.5 &  0.0$\pm$ 4.4 & 345 \\
15 & 15.0$\pm$12.1 & 35.0$\pm$15.5 & 22.5$\pm$13.8 &  2.5$\pm$ 6.5 &  0.0$\pm$ 4.4 & 497 \\
20 &  7.5$\pm$ 9.4 & 10.0$\pm$10.4 & 10.0$\pm$10.4 &  0.0$\pm$ 4.4 &  0.0$\pm$ 4.4 & 636 \\
25 & 20.0$\pm$13.3 & 17.5$\pm$12.7 & 22.5$\pm$13.8 &  0.0$\pm$ 4.4 &  0.0$\pm$ 4.4 & 772 \\
30 & 10.0$\pm$10.4 & 10.0$\pm$10.4 &  7.5$\pm$ 9.4 &  0.0$\pm$ 4.4 &  0.0$\pm$ 4.4 & 912 \\
35 & 15.0$\pm$12.1 & 17.5$\pm$12.7 & 10.0$\pm$10.4 &  0.0$\pm$ 4.4 &  0.0$\pm$ 4.4 & 1033 \\
40 &  7.5$\pm$ 9.4 &  7.5$\pm$ 9.4 & 10.0$\pm$10.4 &  0.0$\pm$ 4.4 &  0.0$\pm$ 4.4 & 1204 \\
45 & 15.0$\pm$12.1 &  2.5$\pm$ 6.5 &  2.5$\pm$ 6.5 &  0.0$\pm$ 4.4 &  0.0$\pm$ 4.4 & 1320 \\
50 & 10.0$\pm$10.4 &  0.0$\pm$ 4.4 &  0.0$\pm$ 4.4 &  0.0$\pm$ 4.4 &  0.0$\pm$ 4.4 & 1497 \\
\bottomrule
\end{tabular}
\end{table*}

\begin{table*}[!t]
\centering
\caption{D2 Symbolic Pointers: Acc\% $\pm$ half-CI by depth.}
\label{app:d2_full}
\small
\begin{tabular}{lcccccc}
\toprule
\textbf{N} & Claude & Gemini & DeepSeek & GPT-4o-m & LLaMA & AvgTok \\
\midrule
 5 & 97.5$\pm$ 6.5 & 55.0$\pm$16.1 & 37.5$\pm$15.7 & 12.5$\pm$11.3 & 40.0$\pm$15.9 & $\sim$205 \\
10 & 90.0$\pm$10.4 & 52.5$\pm$16.2 & 30.0$\pm$15.0 &  0.0$\pm$ 4.4 &  2.5$\pm$ 6.5 & $\sim$380 \\
15 & 85.0$\pm$12.1 & 50.0$\pm$16.2 & 45.0$\pm$16.1 &  0.0$\pm$ 4.4 &  0.0$\pm$ 4.4 & $\sim$555 \\
20 & 67.5$\pm$15.3 & 42.5$\pm$16.0 & 52.5$\pm$16.2 &  0.0$\pm$ 4.4 &  0.0$\pm$ 4.4 & $\sim$725 \\
25 & 80.0$\pm$13.3 & 25.0$\pm$14.3 & 30.0$\pm$15.0 &  0.0$\pm$ 4.4 &  0.0$\pm$ 4.4 & $\sim$895 \\
30 & 57.5$\pm$16.0 & 20.0$\pm$13.3 & 35.0$\pm$15.5 &  0.0$\pm$ 4.4 &  0.0$\pm$ 4.4 & $\sim$1053 \\
35 & 70.0$\pm$15.0 & 22.5$\pm$13.8 & 30.0$\pm$15.0 &  0.0$\pm$ 4.4 &  0.0$\pm$ 4.4 & $\sim$1235 \\
40 & 57.5$\pm$16.0 &  5.0$\pm$ 8.2 & 17.5$\pm$12.7 &  0.0$\pm$ 4.4 &  0.0$\pm$ 4.4 & $\sim$1410 \\
45 & 52.5$\pm$16.2 & 10.0$\pm$10.4 & 22.5$\pm$13.8 &  0.0$\pm$ 4.4 &  0.0$\pm$ 4.4 & $\sim$1580 \\
50 & 55.0$\pm$16.1 & 17.5$\pm$12.7 & 12.5$\pm$11.3 &  0.0$\pm$ 4.4 &  0.0$\pm$ 4.4 & $\sim$1755 \\
\bottomrule
\end{tabular}
\end{table*}

\begin{table*}[!t]
\centering
\caption{D3 Social Logic: Acc\% $\pm$ half-CI by depth.}
\label{app:d3_full}
\small
\begin{tabular}{lcccccc}
\toprule
\textbf{N} & Claude & Gemini & DeepSeek & GPT-4o-m & LLaMA & AvgTok \\
\midrule
 5 & 35.0$\pm$15.5 & 32.5$\pm$15.3 & 20.0$\pm$13.3 &  2.5$\pm$ 6.5 &  0.0$\pm$ 4.4 & $\sim$130 \\
10 &  5.0$\pm$ 8.2 &  0.0$\pm$ 4.4 &  0.0$\pm$ 4.4 &  0.0$\pm$ 4.4 &  0.0$\pm$ 4.4 & $\sim$390 \\
15 &  0.0$\pm$ 4.4 &  0.0$\pm$ 4.4 &  0.0$\pm$ 4.4 &  0.0$\pm$ 4.4 &  0.0$\pm$ 4.4 & $\sim$735 \\
20 &  0.0$\pm$ 4.4 &  0.0$\pm$ 4.4 &  0.0$\pm$ 4.4 &  0.0$\pm$ 4.4 &  0.0$\pm$ 4.4 & $\sim$1265 \\
25 &  5.0$\pm$ 8.2 &  0.0$\pm$ 4.4 &  0.0$\pm$ 4.4 &  0.0$\pm$ 4.4 &  0.0$\pm$ 4.4 & $\sim$1730 \\
30 &  7.5$\pm$ 9.4 &  0.0$\pm$ 4.4 &  0.0$\pm$ 4.4 &  0.0$\pm$ 4.4 &  0.0$\pm$ 4.4 & $\sim$2170 \\
35 &  2.5$\pm$ 6.5 &  0.0$\pm$ 4.4 &  0.0$\pm$ 4.4 &  0.0$\pm$ 4.4 &  0.0$\pm$ 4.4 & $\sim$2780 \\
40 &  2.5$\pm$ 6.5 &  0.0$\pm$ 4.4 &  0.0$\pm$ 4.4 &  0.0$\pm$ 4.4 &  0.0$\pm$ 4.4 & $\sim$3195 \\
45 &  0.0$\pm$ 4.4 &  0.0$\pm$ 4.4 &  0.0$\pm$ 4.4 &  0.0$\pm$ 4.4 &  0.0$\pm$ 4.4 & $\sim$3562 \\
50 &  5.0$\pm$ 8.2 &  0.0$\pm$ 4.4 &  0.0$\pm$ 4.4 &  0.0$\pm$ 4.4 &  0.0$\pm$ 4.4 & $\sim$4317 \\
\bottomrule
\end{tabular}
\end{table*}

\section{Failure Mode Examples and Prompt Templates}
\label{app:examples}

\subsection*{Failure Mode Examples}

\paragraph{D1 early divergence (LLaMA, $N{=}10$).}
\begin{verbatim}
Ground Truth:
  Step 1: [[1,2,3],[4,5,6],[7,8,9]]
  Step 2: [[7,4,1],[8,5,2],[9,6,3]]
Model Output:
  Step 1: [[1,2,3],[4,5,6],[7,8,9]]
  Step 2: [[3,2,1],[6,5,4],[9,8,7]]
            <- DIVERGENCE (k*=2)
\end{verbatim}
Horizontal flip instead of $90^\circ$ CW rotation; $\kstar{=}2$.

\paragraph{D2 constraint failure (Gemini, $N{=}25$).}
After twenty correct steps the model illegally re-assigns variable $A$,
violating uniqueness. Classified Constraint, not Reasoning.

\paragraph{D3 cascade collapse (DeepSeek, $N{=}10$).}
The model processes step 1 correctly but omits a transitive closure
propagation at step 2. Every subsequent pair classification is wrong;
recovery is impossible within the context window.

\subsection*{Exact Prompt Templates}
\label{app:prompts}

\paragraph{D1.}
\begin{verbatim}
You are a spatial reasoning engine.
Track a 3x3 grid
(Initial: [[1,2,3],[4,5,6],[7,8,9]]).
OPERATIONS:
  ROTATE_90_CW: Rotate 90 deg clockwise.
  SHIFT_ROW_2_LEFT: Shift middle row left,
                    wrapping.
Output:
  TRACE: ["Step 1:[[...]]", ...]
  ANSWER: [[...]]
\end{verbatim}

\paragraph{D2.}
\begin{verbatim}
You track 7 variables A-G holding
distinct digits 0-9. Apply N operations:
SHIFT_RIGHT, SET X TO Y PLUS Z mod 10.
Output:
  TRACE: ["Step 1:{A:v,...}", ...]
  ANSWER: {A:v, B:v, ...}
\end{verbatim}

\paragraph{D3 verbose ablation addition.}
\begin{verbatim}
CRITICAL ABLATION INSTRUCTION: After EVERY
single operation, you MUST explicitly
restate the entire agent belief state array
before proceeding to the next step.
\end{verbatim}

\section{Extended Related Work}
\label{app:related_extended}

\paragraph{Depth-scaling and compositional generalisation.}
SCAN \citep{lake2018scan} and BIG-Bench \citep{srivastava2022beyond}
hold difficulty roughly constant and probe systematic generalisation
to novel compositions. \CCB{} complements that line by providing a
continuous, parametric depth axis across three heterogeneous domains
and a single-parameter decay model. SokoBench \citep{sokobench2026}
isolates planning depth in Sokoban; TopoBench
\citep{topobench2026} focuses on topological reasoning and shows
that structured state aids reasoning, motivating tool-augmented D3
extensions.

\paragraph{Relational benchmarks.}
CLUTRR \citep{sinha2019clutrr} tests multi-hop relational reasoning
on kinship graphs and is the closest prior work to D3. \CCB{}
extends that line by providing deterministic ground-truth traces
(not only final answers), enabling TFBC-level diagnostics; by
applying a continuous depth axis from $N{=}5$ to $N{=}50$; and by
integrating relational inference with spatial and symbolic regimes
under a unified evaluation framework.

\paragraph{State tracking and systematic failures.}
\citet{dziri2023faith} showed LLMs unroll memorised subgraphs with
catastrophic failure at compositional OOD depths.
\citet{hou2026wmfam} showed performance degrades under cumulative
state-tracking load. \CCB{} quantifies these phenomena via the
$\kstar$ distribution and the $\pd$ summary statistic.

\paragraph{Trace-level evaluation and structural uncertainty.}
\citet{golovneva2022roscoe} and \citet{prasad2023receval} evaluate
reasoning chains for correctness and informativeness; MME-CoT
\citep{jiang2025mmecot} introduces precision and recall metrics for
multimodal chain-of-thought. \citet{chaudhury2026structural} show
that unstable self-preference rankings signal unreliable inference.
\CCB{} provides a complementary, ground-truth-grounded
operationalisation that requires no LLM-as-judge.

\paragraph{Process supervision and long-horizon execution.}
Process-supervised models \citep{cobbe2021training} are trained with
step-level reward signals that incentivise intermediate-state
correctness; their evaluation is the most consequential extension
of this work. \citet{sinha2025illusion} analytically links per-step
accuracy to an effective task horizon
$H_s{\approx}\ln(s)/\ln(\pd)$; \CCB{}'s empirical $\pd$ values feed
directly into that framework. Recursive scaffolds
\citep{yang2026recursive} and fast-slow recurrence
\citep{takashiro2026thinking} target the same state-management
bottleneck from the architecture side, and \citet{kim2024token} argue
that autoregressive token ordering is itself an inductive bias on
accessible reasoning patterns.

\end{document}